\documentclass[11pt,leqno]{article}
\usepackage{lingmacros}
\usepackage[english]{babel}
\usepackage{chicago}
\usepackage{fol}

\mathchardef\myand="2026

\title{Abductive reasoning with temporal information}
\author{\mbox{\shortstack{Sven Verdoolaege$^1$, Marc Denecker$^1$ and Frank Van Eynde$^2$}}\\
$^1$Department of Computer Science and \\
$^2$Center for
Computational Linguistics\\
K.U.Leuven}
\date{}

\begin{document}
\maketitle

\begin{abstract}
Texts in natural language contain a lot of temporal information,
both explicit and implicit.
Verbs and temporal adjuncts carry most of the explicit information,
but for a full understanding general world knowledge and default
assumptions have to be taken into account.
We will present a theory for
describing the relation between, on the one hand, verbs, their tenses and
adjuncts and, on the other, the eventualities and periods 
of time they represent
and their relative temporal locations,
while allowing interaction with general world knowledge.

The theory is formulated in an extension of first order logic 
and is a practical implementation of the concepts described
in 
\citeN{VanEynde2001}
and 
\shortciteN{Schelkens2000}.
We will show how an
abductive resolution procedure can be used on this representation to
extract temporal information from texts.
The theory presented here is an extension of 
that in \shortciteN{Verdoolaege2000}, adapted to \citeN{VanEynde2001},
with a simplified and extended analysis of adjuncts and with more
emphasis on how a model can be constructed.
\end{abstract}

\section{Introduction}

This article presents some work conducted in the framework of
Linguaduct, a project on the temporal interpretation of Dutch texts by means
of abductive reasoning.

A natural language text contains a lot of both explicit and implicit
temporal information, mostly in verbs and adjuncts. 
The purpose of the theory presented here
is to represent how this information is available in Dutch texts
with the aim of allowing extraction of that information from particular
texts.%
\footnote{In this paper, we only deal with sentences.}
The extracted information contains the temporal relations
between the eventualities described in the text as well as 
relations to periods of time explicitly or implicitly described
in the text. To arrive at this information some (limited)
reasoning needs to be performed on the representation.

The theory is an adaptation of the theory described in 
\citeN{VanEynde1998}, 
\citeN{VanEynde2000} 
and \citeN{VanEynde2001},
which integrates a DRT-style analysis~\cite{Kamp1993} of tense and aspect
into HPSG~\cite{Pollard1994}.
We refer to \citeN{VanEynde2001} for the linguistic terminology used
in this paper.
The adaptation concerns a reformulation of the typed feature
structures in terms of first order logic. This
facilitates the modeling of the interaction between
linguistic knowledge and world knowledge.

The extraction of temporal information is performed by constructing
a model of the logical theory, for which we use an existing
abductive procedure. In case of ambiguities, several models exist
and each can be generated.

We will first describe the knowledge representation language
that we are going to use (essentially first order logic), 
followed by a brief summary
of the representation of the linguistic theory in logic.
Then, we present a short description of the reasoning procedure
used and we finish off with some examples of how it can be used on the theory
to extract temporal information.

\section{Representation language}
\label{s:language}

As already mentioned, the theory is represented in
a language that is essentially first order logic (FOL).
It is extended with some axioms and notational conveniences,
which will be explained in this section.

First of all, we want different constants to represent
different entities and, more generally, we want different
functors and functors with differing arguments to represent
different entities. This means that constants and functors
identify objects; functors can be seen as constructors.
To accomplish this, the so-called
Unique Names Axioms (UNA) are added to the theory \cite{Reiter80a}
\cite{Clark78}.
Since there are an infinite number of such axioms,
they are (implicitly) built into the solver.

Sometimes, however, we want to use open functions,
that is, functions that do not identify an object, but
rather whose result can be equal to the result of another
function. 
The particular solver we use, assumes UNA for every functor,
so we cannot represent an $n$-ary
open function with a functor.
Instead we can use an $(n+1)$-ary predicate, with the extra
argument representing the result of the function and with
an axiom ensuring the existence and uniqueness of the function result
in function of its arguments.
For example, a binary function mapping a country and a year
to the person that is or was the president of that country
in that year, would be represented by a ternary predicate,
e.g.\ $\pres/({\it USA}/,2000,{\it Bill}/)$.

We use a special notation for open functions that
also expresses that the arguments and the result
each satisfy a predicate. For example, the \pres/ function
mapping a country and a year to a person is represented
as follows, where the {\tt{of}}\ indicates the introduction
of an open function.
\begin{equation}
\fol{of pres:: country(_), year(_) -> person(_).}
\end{equation}
The above declaration is equivalent to a set of axioms,
but it is shorter 
and easier to understand and it
allows reasoning procedures working on a
specification to handle such constraints more efficiently.
The following is the more long-winded version:
\begin{equation}
{\obeylines\fol{%
fol forall(C,Y)$ country(C) & year(Y) 
    => exists(P)$ person(P) & pres(C,Y,P).
fol forall(C,Y,P1,P2)$ pres(C,Y,P1) & pres(C,Y,P2) => P1=P2.
fol forall(C,Y,P)$ pres(C,Y,P) => country(C) & year(Y).
}}
\end{equation}
The {\tt fol} marker indicates that what follows is a first
order logic formula.
These axioms express,
first, that for each country and each year, there is a person that
is the president of that country in that year, i.e.~\pres/ is a
total function.
Second, that for a given country and year, there is at most one
president of that country in that year.
Third, that for each instantiation of the president predicate,
the first two arguments are a country and a year
respectively.

While FOL is ideally suited to represent assertional
knowledge, that is, facts and axioms, it does not fare
so well when it comes to definitional knowledge, i.e.\ to defining concepts.
A definition of a concept is an exhaustive enumeration of the
cases in which some object belongs to the concept.
We use a notation borrowed from logic programming.
For example, the concept {\it uncle} is defined as either 
a male sibling (i.e.\ brother) of a parent or a
male spouse of a (presumably female) sibling of a parent:
\begin{equation}
{\obeylines\fol{%
uncle(S,C) <- male(S) & sibling(S,P) & parent(P,C).

uncle(S,C) <- male(S) & married(S,A) & 
sibling(A,P) & parent(P,C).
}}
\end{equation}
Predicates that are not defined are called open predicates.
Open functions are always open predicates.

For the simple, non-recursive, definitions
we use in our theory, such a definition is equivalent to
its completion \cite{Clark78}.\footnote{The equivalence also
holds for a certain subset of recursive definitions.}
The completion of a definition states that 
a predicate defined in it,
holds if and only if one of its cases holds.
That is, the defined predicate, which is the head of
each rule, is equivalent to
the disjunction of the bodies of the rules.
In order to be able to take the disjunction of the bodies,
the heads have to be identical, of course.
To accomplish this, all terms
in an argument position of the predicate
in the head are first moved to the body as an equality to the variable
representing that argument, and all variables local to the
body are quantified in the body.
That is the set of
$m$ rules defining $p/n$: $\forall x_i:p(t_i)\leftarrow F_i$
is turned into the following equivalence:
\begin{equation}
\forall{\bf z}: p({\bf z}) \leftrightarrow \left \{ \begin{array}{c}
(\exists x_1: {\bf z=t}_1 \land F_1) \\
\lor\\
\ldots\\
\lor\\
(\exists x_m:\  {\bf z=t}_m \land F_m)
\end{array}\right.
\end{equation}
\citeN{Denecker2000c} presents the extension of
classical logic with a more general notion of (inductive)
definitions.

\section{Representation}

The way in which information can be extracted from a theory
is of course largely dependent on how the theory is represented.
We will therefore first discuss this representation and only
in the following sections will we present the extraction part.

\def\lb#1{$_{\hbox{\tiny #1}}$}

This section will mainly focus on the sentence ``[\lb{S} [\lb{NP} Ik] 
[\lb{VP} [\lb{V-AUX} ben] [\lb{VP} [\lb{ADV} gisteren]
[\lb{VP} [\lb{ADJ} ziek] [\lb{V-MAIN} geweest]]]]]'' 
(I have been sick yesterday),
showing how this sentence
is represented and showing the parts of the theory needed to extract
information from it.%
\footnote{Note that in general, there is no direct correspondence
between Dutch and English tenses.} 

\subsection{Input}
\label{s:input}
The sentence ``Ik ben gisteren ziek geweest'' contains three
interesting words when it comes to temporal information:
``ben'', ``geweest'' and ``gisteren''. The first
two are both forms of the verb ``zijn'' (to be), the last
is a temporal adjunct.

To be able to refer to different occurrences of the
same word (such as ``zijn'' in the example,
which occurs both in its past participle form
and in its present tense form), we make
use of tokens which are arbitrarily chosen constants
that represent these occurrences. In our example sentence
({\it s1}), we have verb tokens {\it w1} for the main
verb and {\it w2} for the auxiliary verb, and an adjunct
token {\it a1}. 

To express which word is associated with a token, we use
the \verbtword/\ and \adjtword/\ predicates. For verbs,
we additionally use the \vform/\ predicate to indicate the
verb form. Since all of these are exhaustive enumerations,
they are defined predicates.
For our example sentence we have:
\begin{equation}
\label{e:input}
{\obeylines\fol{%
verbt_word(w1,zijn) <- true.
verbt_word(w2,zijn) <- true.
adjt_word(a1,gisteren) <- true.
vform(w1,past_participle) <- true.
vform(w2,present_tense) <- true.
}}
\end{equation}

To express further that {\it s1} is a clause, 
that {\it w1} is the main verb of {\it s1},
that {\it w2} is an auxiliary verb with {\it w1} as its
complement and that {\it a1} is an adjunct in {\it s1},
we use the following:
\begin{equation}
\label{e:input2}
{\obeylines\fol{%
clause(s1) <- true.
main_verb(s1,w1) <- true.
aux_verb(w2,w1) <- true.
s_adjunct(s1,a1) <- true.
}}
\end{equation}
Here again, we are dealing with exhaustive enumerations and thus
definitions.

\subsection{Periods of time}

The allowed kinds of periods of time are (half-open) intervals, denoted
by the \int/ predicate, and points, denoted by the \point/ predicate.
Intervals refer directly to the time axis.
They are represented by a pair of points on the time axis
as arguments to an \int/ functor. This does not preclude
intervals from having unknown end points. The points on the time
axis themselves can be left unspecified.
Each point on the axis is represented by a function (\ts/) of
four integers, the year, the month, the day of the month
and the hour.\footnote{The rather low resolution is due to
technical limitations.} For example, the whole of May the 21st 1976
is represented as $\int/(\ts/(1976,5,21,0),\ts/(1976,5,22,0))$.

The relations that can be specified between intervals 
deviate from those
in \citeN{Allen83} since we use half-open intervals
and we only need a specific subset of them.
The \overlap/ relation indicates a non-empty intersection,
\within/: an inclusion of the first argument within the second, 
\before/: a precedence and \meets/: an immediate precedence.
These relations 
and some other properties about intervals that we will see later
(e.g.\ \daya/\ and \hour/) are defined in terms of relations and
properties of these end points, which are in turn
defined in terms of relations on their composing parts.
We will not show them here.

\subsection{Tenses and auxiliaries}
\label{s:tense}

This section presents the transformation of some of the rules
in~\cite{VanEynde2001} and is necessarily brief. For a slightly
more detailed exposition, we refer to \shortciteN{Verdoolaege2000},
which also explains our use of the temporal perspective time.
In this paper, we make abstraction of the difference between
the temporal perspective time and the utterance time.

The Dutch language has two simple tenses, the simple present and 
the simple past,
and several others that require auxiliaries.
To indicate which verbs are auxiliaries and if so what
kind of auxiliaries (e.g.\ perfect, future), we use
the \verbauxkind/\ predicate.
For example, 
``zijn'' has (at least)
three uses: one as the main verb of a clause (\vzijn/, such as {\it w1}), 
one as a temporal auxiliary 
(\tzijn/, such as {\it w2}) and one as an aspectual auxiliary (\azijn/).
Both uses of ``zijn'' as an auxiliary are auxiliaries of
the perfect.
\begin{equation}
\label{e:auxlex}
{\obeylines\fol{%
verb_aux_kind(t_zijn,perfect) <- true .
verb_aux_kind(a_zijn,perfect) <- true .
}}
\end{equation}
Each meaning of ``zijn'' refers to the same verb lexeme.
The \verblex/ predicate enumerates the lexemes and we naturally only
show part of its definition here.
\begin{equation}
\label{e:zijnlex}
{\obeylines\fol{%
verb_lex(t_zijn,zijn) <- true .
verb_lex(a_zijn,zijn) <- true .
verb_lex(v_zijn,zijn) <- true .
}}
\end{equation}
It determines
of which verb a verb token can be an occurrence,
by placing a constraint on the open function \tokenverb/ that
maps verb tokens to their corresponding verbs.
The constraint is that the verb is one of the possible
meanings of the word associated with the verb token,
i.e.~that the word of the verb token is the same
as the word of the verb.
\begin{equation}
\label{e:tokenverb}
{\obeylines\fol{%
of token_verb:: verb_token(_) -> verb(_).
fol forall(T,V,W,L)$ token_verb(T,V) & 
verbt_word(T,W) & verb_lex(V,L) => W=L.
}}
\end{equation}
The \verbtoken/ and \verb/ predicates merely
enumerate the verb tokens and verbs respectively.

For each predicate representing one of the other properties
of verbs (such as \verbauxkind/ above), 
there is a corresponding predicate for verb tokens,
that is defined to be that of the verb for which it is a token.
The predicates have the same name, with an extra {\it verb\_}
prefix for the one on verbs.
For example:
\begin{equation}
{\obeylines\fol{%
aux_kind(W) <- exists(V)$ token_verb(W,V) & verb_aux_kind(V).
}}
\end{equation}

For \auxkind/, we need to place an extra (indirect) constraint
on the \tokenverb/\ predicate.
The tokens that appear as an auxiliary in the sentence
(i.e.\ $\exists(\mbox{\it W2}) : \auxverb/(W,\mbox{\it W2})$)
should be precisely those whose verb is an auxiliary
(i.e.\ $\exists(F): \auxkind/(W,F)$).
\begin{equation}
{\obeylines\catcode`\$=12\fol{%
fol forall(W)$ (exists(W2)$ aux_verb(W,W2)) <=> (exists(F)$ aux_kind(W,F)).
}}
\end{equation}

Substantivity is another one of those properties on verbs (\verbsubst/)
that is inherited by verb tokens (\subst/).
Due to the low number of vacuous (i.e.~non-substantive) verbs,
it is simpler to list those instead of all the substantive ones
and then to define the substantive verbs as those
that are not vacuous.
We only show part of the enumeration of vacuous verbs.
\begin{equation}
{\obeylines\fol{%
verb_subst(V) <- not verb_vacuous(V).
}}
\end{equation}
\begin{equation}
\label{e:vacuous}
{\obeylines\fol{%
verb_vacuous(t_hebben) <- true.
verb_vacuous(t_zijn) <- true. 
verb_vacuous(t_zullen) <- true.
}}
\end{equation}

Substantivity determines which verb tokens are associated
with an eventuality. The eventuality is indicated by the unary
\evt/ functor and an additional \isevt/ predicate indicates
which \evt/\,s represent eventualities.
\begin{equation}
\label{e:isevt}
{\obeylines\fol{%
isevt(evt(W)) <- subst(W).
}}
\end{equation}

Each eventuality has both an situation time and a location time
associated with it. 
The situation time is expressed
through the \sittime/ predicate. Similarly, the location time
uses the \loc/\ predicate. The second argument of both is an
interval and is required to be unique for a given eventuality.
This can trivially be expressed by a set of axioms that have
been omitted here. In general, an overlap relation holds between
the two (expressed in the first of the following axioms).
The second axiom expresses
the fact that in case the location time is bounded, 
this overlap is narrowed down to inclusion.
\begin{equation}
\label{e:icevent}
{\obeylines\fol{%
fol forall(E)$ isevt(E)  => 
(exists(L,T)$ loc(E,L)& sittime(E,T)& overlap(T,L)).
fol forall(E,L)$ loc(E,L) & bounded(L) => 
(exists(T)$ sittime(E,T)&within(T,L)).
}}
\end{equation}

One way for the location time to be bounded is if it
belongs to a past participle.
\begin{equation}
\label{e:psp}
{\obeylines\fol{%
fol forall(W)$ vform(W,past_participle)                          
        => (exists(L)$ subst(W) & loc(evt(W),L) & bounded(L)).
}}
\end{equation}

The utterance time is expressed through the unary \utt/\ predicate
and is required to exist.
Essentially, it is just the situation time of some ``\utt/'' eventuality.
\begin{equation}
\label{e:utt}
{\obeylines\fol{%
fol exists(U)$ utt(U).
utt(U) <- sittime(utt,U).
}}
\end{equation}

The linguistic theory specifies three kinds of relations
between the location time and the temporal perspective time,
namely: precedence (\before/) for simple past and temporal
perfect, non-precedence (\notbefore/) for the simple present
and succession (\after/) for the future.
This results in the following definition of \locppp/.
The axioms that actually exert the relations are straight-forward
and have been omitted.
\begin{equation}
\label{e:locppp}
{\obeylines\fol{%
loc_ppp(W,before)<- subst(W) & vform(W,past_tense);
            (exists(A)$ aux_kind(A,perfect) & 
	    not subst(A) &  aux_verb(A,W)).
loc_ppp(W,not_before) <- subst(W) & vform(W,present_tense).
loc_ppp(W,after) <- exists(A)$ aux_kind(A,futurate) & 
	    aux_verb(A,W).
}}
\end{equation}
As you can see, for each clause there is exactly one eventuality
whose location time is related to the temporal perspective time,
either that of the tensed verb in case it is substantive or that
of the complement of the (vacuous) temporal auxiliary.
The above definition is simplified to not deal with transposition.
As such it also does not deal with past tense auxiliaries of
the perfect, since they require transposition.

The rule for the aspectual perfect, shown below, shows that
the auxiliary ({\it w2})
in our example sentence can only be interpreted as a temporal auxiliary.
\begin{equation}
\label{e:aspectual}
{\obeylines\fol{%
fol forall(W,C)$ aux_kind(W,perfect) & subst(W) & aux_verb(W,C)
        => not stative(evt(C)) & result(evt(C),evt(W)).
}}
\end{equation}
As already explained, the axiom states that the aspectual (substantive)
perfect introduces the resulting state of the eventuality of its
(non-stative) complement.

The \result/\ predicate in the above axiom is an open predicate
expressing that the second argument represents the resulting
state of the first argument. This implies that the resulting
state immediately follows the eventuality that it is a resulting
state for.
\begin{equation}
\label{e:result}
{\obeylines\fol{%
fol forall(E,E2,L,L2)$ result(E,E2) & sittime(E,L) & 
        sittime(E2,L2) => meets(L,L2).
}}
\end{equation}

\subsection{Adjuncts}
\label{s:adjuncts}

With respect to adjuncts, our theory is mainly based on 
\shortciteN{Schelkens2000}. We 
distinguish
 between
{\em frame} or {\em locating adjuncts} indicating
when an eventuality occurs, and {\em durational
adjuncts} expressing how long it takes.%
\footnote{We do not deal with frequency adjuncts in this paper.}
Amongst the frame adjuncts, we further 
distinguish
between {\em deictic} adjuncts that refer to the
utterance time (or more generally, to the temporal perspective time)
and {\em independent} adjuncts that refer directly
to the time axis.
In this paper, we do not deal with the so-called anaphoric
adjuncts.

We associate with each adjunct token the period of time that
it refers to. 
For each adjunct, there is an axiom that specifies
or somehow constrains this period of time.
For durational adjuncts, which we will not
discuss further in this paper, only the length of the
period of time is constrained by the adjunct.

For example, the adjunct ``gisteren'' (yesterday),
is a frame adjunct that refers to the day before
the day that includes the temporal perspective time.
\begin{equation}
\label{e:gisteren}
{\obeylines\fol{%
fol forall(A,Y,P)$ 
adjt_word(A,gisteren) & adjtime(A,Y) & adjt_ppp(A,P) => 
 (exists(T)$ day_a(Y), day_a(T), within(P,T) & meets(Y,T)).
}}
\end{equation}
Here, \adjtime/ is another open function linking an adjunct 
to its associated period of time
and \adjtppp/ does the same for the temporal perspective time.
The first is an open function from ``adjunctoids'', which refers to anything
that occurs as a first argument to the \adjtword/ predicate,
to intervals. The second is defined as the temporal perspective time
of the modified verb.
We will show the use of adjunctoids later in this subsection.

\begin{equation}
{\obeylines\fol{%
adjunctoid(A) <- exists(L)$ adjt_word(A,L).
of adjtime:: adjunctoid(_) -> int(_).
}}
\end{equation}

The effect of a frame adjunct is that the location time of the
modified verb is required to be within the frame period:
\begin{equation}
\label{e:frame}
{\obeylines\fol{%
fol forall(A,T,W,L)$  adjunct_verb(A,W) & frame(A) & 
  adjtime(A,T) & loc(evt(W),L) =>             within(L,T).                                                           
}}
\end{equation}
The \adjunctverb/ predicate is an open function that maps adjuncts (anything
that appears as a second argument to \sadjunct/) to one of the
substantive verbs in the clause. This mapping is not specified
in the input, since in general, when a clause contains more
than one (possibly) substantive verb, we do not
know a priori which of these verbs is modified by the adjunct.
The {\it frame} predicate lists all the frame adjuncts. Part of its
definition is as follows:
\begin{equation}
\fol{frame(A) <- adjt_word(A,gisteren).}
\end{equation}

This setting also allows for some more complicated uses,
for example, {\em na} (after) followed by something that
could be used as a frame adjunct. The use of {\em na} 
indicates that the location time of the modified verb
is situated after the period of time of the complement
of the preposition.
This can 
be modeled as inclusion in a frame interval that immediately
succeeds the frame interval of the complement. 
\begin{equation}
{\obeylines\fol{%
fol forall(A,B,T,F,P)$ adjt_word(A,na(B)) & adjtime(A,T) & 
frame(B) & adjtime(B,F) => meets(F,T)
}}
\end{equation}
The definition of {\it frame} is extended accordingly:
\begin{equation}
{\obeylines\fol{%
frame(A) <- adjt_word(A,na(B)) & frame(B).
}}
\end{equation}
In the input, {\em na gisteren} would be specified as follows:
\begin{equation}
{\obeylines\fol{%
adjt_word(a1,gisteren) <- true.
adjt_word(a2,na(a1)) <- true.
s_adjunct(s1,a2) <- true.
}}
\end{equation}

\shortciteN{Schelkens2000} also discusses so-called {\em point-like}
frame adjuncts such as ``om x uur'' (at x o'clock) and argues
that they affect both the location time and the situation time,
in that the point described by the adjunct in included in both.
As this is different from the behaviour of frame adjuncts belonging
to the \frame/ predicate, we introduce another one, \pointframe/,
which lists all the point-like frame adjuncts. 
The effect these have is described by the following axiom.

\begin{equation}
\label{e:pointframe}
{\obeylines\fol{%
fol forall(A,T,W,L,E)$                                                          
     point_frame(A) & adjtime(A,T) & verb_adjunct(W,A) & 
    loc(evt(W),L) &        sittime(evt(W),E) =>                                                       
         within(T,L) & within(T,E).                                             
}}
\end{equation}

\section{Reasoning procedure}
\label{s:procedure}

Up to this point, we have presented a logic theory consisting
of FOL axioms and definitions in which some predicates were
defined and others open, but we have not shown how to derive
information from this theory. Since the defined predicates
are known, the only uncertainty lies in the open predicates.
We need to find an interpretation for these open predicates
that is consistent with the theory. In other words,
we need to generate a model for the open predicates.
To do this we use an existing abductive procedure,
called SLDNFA \cite{VanNuffelen99}, which operates on theories written in
the knowledge representation language outlined in 
section~\ref{s:language}.

Abduction is a form of non-monotonic reasoning that
is used to explain observations. In this case,
an explanation consists of a model for the open predicates.
To explain some observation, we sometimes have to
assume (abduce) other information.
This form of reasoning is called non-monotonic, because new
information may invalidate previously drawn conclusions.

We cannot give a detailed explanation of how the SLDNFA procedure
works within this limited space, but we do want to give an idea of 
it.
In short, the procedure tries to make the conjunction of
all axioms and the (possibly empty) observation (query) hold
by abducing a set of atoms, according to the following rules. 
A conjunction holds if all of its
conjuncts hold; for a disjunction, it is sufficient that one of
its disjuncts holds, so each one is tried out until one is found
that holds. Negation is distributed over disjunctions and conjunctions. 
A defined predicate is replaced by its completion 
(see section~\ref{s:language}).

When an open predicate occurs negatively, the atom is 
(temporarily) assumed not
to hold if this does not conflict with earlier made assumptions.
If it is part of a disjunction, the remaining disjuncts
are remembered in case it turns out we want it to hold after all.
When an open predicate occurs positively, the atom is assumed to
hold and any applicable remembered disjunction is required
to hold as well. If the atom was already assumed not to hold
and there were no remaining disjuncts at that point,
the procedure backtracks to the latest disjunction with remaining
disjuncts. Finally, an equality unifies its arguments. If this fails,
the procedure backtracks as well.

Numerical operations and comparisons are
translated into CLP (Constraint Logic
Programming) constraints and handed over to
an efficient constraint solver.
They occur mainly in the definitions and axioms
pertaining to points on the time axis, which we
have not shown in this paper.
At the end, all numerical variables are
labeled, which means that they
get a value assigned to them that satisfies
all constraints.

\section{Deriving temporal information}

Applying the procedure of section~\ref{s:procedure} to our
theory with an empty observation (that is, we just want a
consistent interpretation for the open predicates)
for our example sentence ``Ik ben gisteren ziek geweest'',
we get the following result, which lists for each open predicate
precisely its model:

\begin{verbatim}
adjt_ppp: [adjt_ppp(a1,int(ts(1999,1,2,0),ts(1999,1,2,1)))]
adjtime: [adjtime(a1,int(ts(1999,1,1,0),ts(1999,1,2,0)))]
adjunct_verb: [adjunct_verb(a1,w1)]
sittime: [sittime(evt(w1),int(ts(1999,1,1,0),ts(1999,1,2,0))),
          sittime(utt,int(ts(1999,1,1,0),ts(1999,1,3,0)))]
loc: [loc(evt(w1),int(ts(1999,1,1,0),ts(1999,1,2,0)))]
s_ppp: [s_ppp(s1,int(ts(1999,1,2,0),ts(1999,1,2,1)))]
token_verb: [token_verb(w2,t_zijn),token_verb(w1,v_zijn)]
\end{verbatim}

We will now show how this result can be obtained from the
given axioms and definitions, without, however, following
the exact procedure as presented in section~\ref{s:procedure}
as that would be too tedious.

The constructed model should satisfy each axiom.
Arguably the simplest axiom is the one requiring the
existence of an utterance time~(\ref{e:utt}), which
requires us to abduce an instance of the \sittime/ predicate,
viz.~ one with {\it utt} as first argument and {\em some} $U$
as second argument. Further constraints will narrow down
the possible values for this $U$.

Next, we look at the correspondence between verb tokens and 
verbs, i.e.~\tokenverb/. 
Equation~(\ref{e:tokenverb}) shows that \tokenverb/ is
a total function from verb tokens to verbs with the
restriction that they refer to the same lexeme.
The input~(\ref{e:input}) shows that both verb tokens ({\it w1}
and {\it w2}) refer to {\it zijn}, for which equation~(\ref{e:zijnlex})
allows three possibilities. 
Now, we know from~(\ref{e:input2}) that {\it w1} is not an auxiliary,
so, based on~(\ref{e:auxlex}), it cannot be \tzijn/ or \azijn/, so
it must be \vzijn/. Similarly,
{\it w2} is an auxiliary
and since the axiom on the aspectual perfect~(\ref{e:aspectual}) requires
non-stativity of the complement, whereas \vzijn/ is stative,
it must be temporal, i.e.~\tzijn/.\footnote{The definition of \stative/
is not shown, but is similar to \subst/ and includes \vzijn/.}
The abductive procedure would abduce each possibility in turn
and would only reconsider when it hits a contradiction such as
mentioned above.

Since \vzijn/ is substantive (it is not listed as vacuous~(\ref{e:vacuous})),
it introduces an eventuality~(\ref{e:isevt}), and therefore it has
overlapping eventuality and location times~(\ref{e:icevent}).
Since {\it w1} is furthermore a past participle, the location time
is bounded~(\ref{e:psp}) and thus even {\em includes} the situation time.
As to the relation with the temporal perspective time, 
the \locppp/ definition~(\ref{e:locppp}) specifies a before relation
for temporal (non substantive) auxiliaries of the perfect.

Finally, we look at the information contained in the adjunct.
As mentioned in section~\ref{s:adjuncts},
\adjunctverb/ associates adjunct tokens to tokens of substantive
verbs of the same clause. In this case there is only one
substantive verb ({\it w1}), so the adjunct can only modify that verb.
Since we are dealing with the adjunct ``gisteren'', 
the right-hand side of implication in axiom~(\ref{e:gisteren}) has to hold,
namely that the adjunct refers to a day that immediately precedes
the day that contains the temporal perspective time.
Axiom~(\ref{e:frame}) then further ensures that this period of time
includes the location time of \evt/({\it w1}).

At this stage, all the axioms are satisfied, but our model is
underspecified in that some abduced atoms still contain variables,
notably those that refer to time intervals.
We can then choose these time intervals, of course taking into
account the constraints that have been placed on them.
In this case, the abductive procedure has chosen
the temporal perspective time to be the first
hour of January the second, 1999. It is included in the utterance
time, which,
as you can see, is rather large, stretching over two days.%
\footnote{Maybe some general restriction
should be placed on the extent of the utterance time.}
The adjunct ``gisteren'' (yesterday) then of course refers
to January the first and this includes the location time of
the eventuality associated with {\it w1} (the whole day, here),
which in turn includes the situation time (also the whole day, here).

When more information is given, the system does not have that
much choice.
Suppose, for example, that you know that the sickness 
lasted from 6 o'clock until
8 o'clock in the evening on May the 21st, 2000 and that the
utterance time lasted an hour, you would give the following
observation (query) to the system:
\begin{equation}
{\obeylines\fol{%
utt(U) &hour(U) &
 sittime(evt(w1),int(ts(2000,5,21,18),ts(2000,5,21,20)))
}}
\end{equation}
The result is, as you would expect, that the utterance happened on
May the 22nd. We only show the result for the \sittime/ predicate;
the other time intervals are changed accordingly.
\begin{verbatim}
sittime: [sittime(utt,int(ts(2000,5,22,0),ts(2000,5,22,1))),
      sittime(evt(w1),int(ts(2000,5,21,18),ts(2000,5,21,20)))]
\end{verbatim}

An example of another kind of query, is the following.
Is it possible for the utterance to have taken place
before the eventuality described by {\it w1} ?
\begin{equation}
{\obeylines\fol{%
utt(U) & sittime(evt(w1),T) & before(U,T)
}}
\end{equation}
The result:
\begin{verbatim}
no
\end{verbatim}

The example sentence used is the previous queries, gave rise
to at most one model (apart from the possible choice of time
periods). 
We can also analyze sentences with ambiguities.
For example, \shortciteN{VanEynde2001} argues that sentence \ref{e:eten},
reproduced below, has two possible interpretations, depending
on which of the verbs is taken to be modified by the adjunct.

\setcounter{enums}{17}
\enumsentence{\label{e:eten}\shortex{9}
    {Ze & had & om & vier & uur & al & een & hamburger & gegeten.}
    {she & have.{\sc past} & at & four & hour & already & a & hamburger & eat.{\sc psp}}
    {`She had already eaten a hamburger at four o'clock.'}
}

The input is a straightforward transliteration:
\begin{equation}
{\obeylines\fol{%
main_verb(s1,w1) <- true.                                                       
aux_verb(w2,w1) <- true.                                                        
verbt_word(w1,eten) <- true.                                                    
verbt_word(w2,hebben) <- true.                                                  
vform(w1,past_participle) <- true.                                              
vform(w2,past_tense) <- true.                                                   
s_adjunct(s1,a1) <- true.                                                       
adjt_word(a1,om(4)) <- true.                                                    
}}
\end{equation}
When given this sentence as input, we do indeed get two
significantly different models, of which we only show
the relevant parts. Both have identified {\em w2}
as an aspectual perfect. According to~(\ref{e:aspectual})
and~(\ref{e:result}), this means that the eventuality of
having eaten ({\em evt(w2)}) immediately 
follows the eventuality of eating ({\em evt(w1)}).
The first model then takes the adjunct to modify the main verb
and (~\ref{e:pointframe}) forces the time referred to
by the adjunct to included in the eating.

\begin{verbatim}
adjtime: [adjtime(a1,int(ts(1999,1,1,4),ts(1999,1,1,5)))]
adjunct_verb: [adjunct_verb(a1,w1)]
sittime: [sittime(evt(w1),int(ts(1999,1,1,0),ts(1999,1,2,0))),
          sittime(evt(w2),int(ts(1999,1,2,0),ts(1999,1,3,0)))]
token_verb: [token_verb(w2,a_hebben),token_verb(w1,v_eten)]
\end{verbatim}

Similarly, in the second model the auxiliary is modified
by the adjunct and it it the having eaten that includes
the time referred to by the adjunct.

\begin{verbatim}
adjtime: [adjtime(a1,int(ts(1999,1,2,4),ts(1999,1,2,5)))]
adjunct_verb: [adjunct_verb(a1,w2)]
sittime: [sittime(evt(w1),int(ts(1999,1,1,0),ts(1999,1,2,0))),
          sittime(evt(w2),int(ts(1999,1,2,0),ts(1999,1,3,0)))]
token_verb: [token_verb(w2,a_hebben),token_verb(w1,v_eten)]
\end{verbatim}

\section{Conclusions}

In this paper, we have shown a formalisation 
in first order logic
of an existing
theory about temporal information in Dutch texts,
that is an extension and enhancement of the formalisation
in \shortciteN{Verdoolaege2000}.

The representation shown in this paper has effectively been
implemented and results of  using an abductive reasoning 
procedure on it were presented.
Although the problems dealt with in this paper were limited,
the experiments show that abduction may be viable for
natural language processing.

A representation in logic is not only very flexible
and extensible, a good representation also requires the use of well
thought out concepts, because of logic's clear
and formal semantics.
This representation has already helped in clearing out
the meaning of some concepts used in the temporal analysis
of sentences.

The research that led to this paper has also been
an exercise in knowledge representation and will
contribute toward a better knowledge representation
methodology.

\bibliography{iwcs4}
\bibliographystyle{chicago}

\end{document}